\documentclass[11pt]{article}
\usepackage[utf8]{inputenc}
\usepackage{float}

\usepackage[preprint]{acl}

\usepackage{times}
\usepackage{latexsym}
\usepackage{tabularx}
\usepackage[T1]{fontenc}

\usepackage[utf8]{inputenc}

\DeclareUnicodeCharacter{258E}{}
\usepackage{microtype}

\usepackage[scaled=0.92]{inconsolata}

\usepackage{graphicx}
\usepackage{amsmath}
\usepackage{enumitem}
\usepackage{booktabs}
\usepackage{multirow}
\usepackage{algorithm}
\usepackage{algorithmic}
\usepackage{balance}
  \usepackage{tcolorbox}
  \tcbuselibrary{breakable}
\usepackage{fancyvrb}
\usepackage{fvextra}

\newtcolorbox{promptbox}[1][]{
    breakable,
    colback=gray!8,
    colframe=gray!50,
    fonttitle=\bfseries\small,
    title=#1,
    left=6pt, right=6pt, top=6pt, bottom=6pt,
    fontupper=\small,
    before upper={\setlength{\parskip}{4pt}}
  }

\usepackage{tikz}
\usepackage{pgfplots}
\pgfplotsset{compat=1.18}

\usepackage{xcolor}

\title{Tree-of-Ideas: Automated Research Ideation via Cross-Trajectory Reasoning over Scholarly Evolution}

\author{
\textbf{Xun Li\textsuperscript{1,2}},
\textbf{Yiying Yang\textsuperscript{1}},
\textbf{Pengtao Li\textsuperscript{3}},
\textbf{Xiao Yao\textsuperscript{1,4}},
\textbf{Suyu Liu\textsuperscript{5}},
\textbf{Xiaoyang Ye\textsuperscript{1,4}},
\\
\textbf{Ziyu Lu\textsuperscript{1,4}},
\textbf{Yuan Yao\textsuperscript{1}},
\textbf{Yangning Li\textsuperscript{6}},
\textbf{Yinghui Li\textsuperscript{7}},
\textbf{Wenhao Jiang\textsuperscript{1,*}}
\\[0.5em]
{\footnotesize \textsuperscript{1}Guangdong Laboratory of Artificial Intelligence and Digital Economy (SZ)
\quad
\textsuperscript{2}University of Melbourne}
\\[-0.05em]
{\footnotesize \textsuperscript{3}University of Sydney
\quad
\textsuperscript{4}Shenzhen University
\quad
\textsuperscript{5}Nanyang Technological University}
\\[-0.05em]
{\footnotesize \textsuperscript{6}Tsinghua University
\quad
\textsuperscript{7}Tencent Youtu Lab, Tencent}
\\[0.15em]
{\footnotesize
\textsuperscript{*}\textbf{Corresponding author:}
\href{mailto:cswhjiang@gmail.com}{\texttt{cswhjiang@gmail.com}}
}
}

\begin{document}
\maketitle
\begin{abstract}
Effective research ideation requires moving beyond a static understanding of prior work to trace how research problems and solutions evolve across the literature. Existing methods either treat papers as unstructured context or model scholarly evolution as isolated citation chains, overlooking interactions among research trajectories. We propose  \textbf{Tree-of-Ideas} (ToI), a two-stage framework. \textbf{EvoTrace} reconstructs branching scholarly trajectories from citations, tracking evolving methods, resolved problems, and gaps. \textbf{EvoAgent} then reasons across trajectories to identify convergent problems and complementary solutions, generating grounded research ideas. Across six AI research topics, ToI achieves the highest score among automatic methods ($6.27$ vs. $5.36$ for the strongest baseline on a 10-point scale), with strong \emph{Novelty} ($6.36$) and \emph{Groundedness} ($7.00$). Also, its score approaches that of human-paper references ($6.29$), demonstrating the value of cross-path evolutionary reasoning.

\end{abstract}
\section{Introduction}
\label{sec:intro}
Most research ideas emerge from understanding how a line of work changes over time. Later studies inherit assumptions and mechanisms from earlier work, modify them to address particular limitations, and often introduce new gaps in the process. Research ideation can therefore be viewed as an evolutionary reasoning problem: a useful system should identify not only which papers are relevant, but also how problems, methods, and unresolved questions develop across the literature. Moreover, some opportunities become visible only when multiple research trajectories are considered together.

Recent LLM-based ideation systems have made substantial progress, yet their representations of prior work are often either too flat or too linear. Retrieval-based systems provide relevant papers or summaries as context, but typically leave the relationships among them implicit~\citep{baek-etal-2024-researchagent,lu-etal-2024-aiscientist}.
Chain-based systems organize papers into chronological or citation trajectories, but represent a branching research landscape through isolated linear paths~\citep{li-etal-2024-chain}. Even when citation graphs are available, they are often used primarily to organize or retrieve papers rather than to reason about how knowledge changes across citation relations. Consequently, these approaches provide limited support for identifying opportunities that arise from interactions among multiple research trajectories.

Addressing this limitation requires reasoning about scholarly evolution at two complementary levels. Within each trajectory, an ideation system should trace how a method advances prior work, what limitation it addresses, and which gap remains or subsequently emerges. Across trajectories, it should identify recurring gaps and complementary mechanisms—for example, when independently evolving branches converge on a related problem, or when a mechanism from one branch can address a gap in another. Such opportunities are difficult to recover from unordered literature context or isolated linear trajectories.

\begin{figure*}[t]
  \centering
  \includegraphics[width=\textwidth]{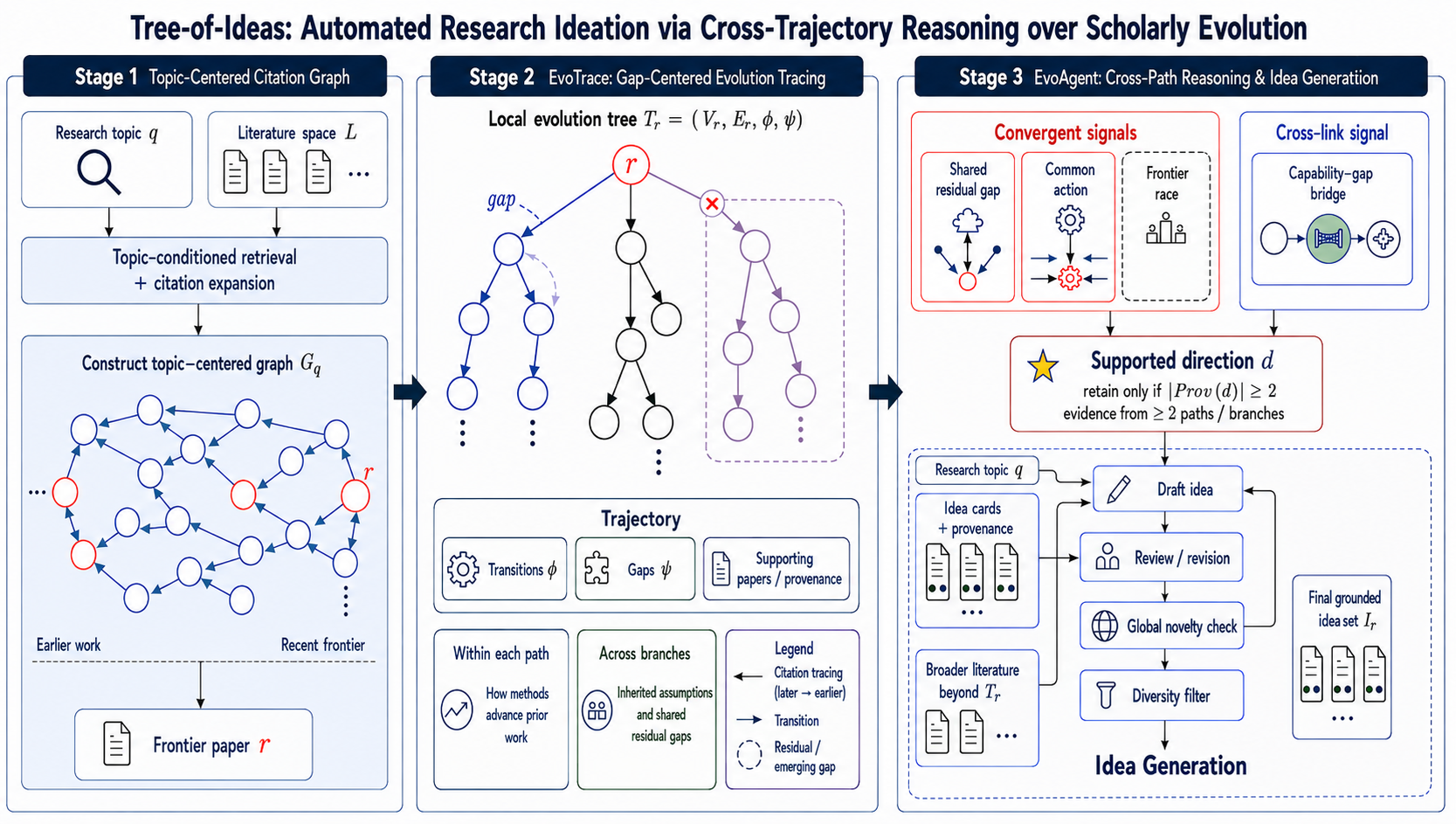}
  \caption{Overview of the ToI framework. Given a research topic and a literature corpus, EvoGraph retrieves and expands relevant papers to construct a topic-centered citation graph and identify frontier papers (Stage~1). EvoTrace reconstructs a local evolution tree for each frontier paper, tracing methodological transitions, unresolved research gaps, and supporting provenance within and across branches (Stage~2). EvoAgent extracts convergent and cross-link signals, retains directions supported by multiple trajectories, and iteratively generates, reviews, and filters ideas to produce a diverse set of literature-grounded research proposals (Stage~3).}
  \label{fig:framework}
\end{figure*}

To this end, we propose \textbf{Tree-of-Ideas} (ToI), an automated research ideation framework that reasons over branching scholarly trajectories, as shown in Figure~\ref{fig:framework}. ToI comprises a topic-centered graph construction stage followed by two coordinated reasoning stages: within-trajectory evolution tracing (\textbf{EvoTrace}) and cross-trajectory ideation (\textbf{EvoAgent}). Given a research topic, ToI first constructs a citation graph and identifies frontier papers as entry points for trajectory reconstruction. \textbf{EvoTrace} then recursively follows citation relations from these papers, tracing how methods advance prior work, which limitations they address, and which gaps emerge or remain unresolved. Finally, \textbf{EvoAgent} reasons across the reconstructed trajectories to identify convergent patterns shared by multiple paths and cross-link opportunities where a mechanism from one path can address a gap in another. It synthesizes these signals into research ideas grounded in explicit multi-path provenance.
Experiments across six AI research topics show that ToI achieves the best overall performance among automatic ideation methods. 

Our contributions are highlighted as follows:
\begin{itemize}[noitemsep, topsep=2pt, leftmargin=1.4em]
\item We present \textbf{Tree-of-Ideas} (ToI), an end-to-end framework for automated research ideation through reasoning over branching scholarly evolution. ToI integrates within-trajectory evolution tracing with cross-trajectory opportunity discovery, moving beyond flat literature contexts and isolated research chains.

\item We develop \textbf{EvoTrace}, which reconstructs gap-centered scholarly evolution trajectories from citation relations. It traces how each work advances its predecessors, which limitations are addressed, and, critically, which gaps emerge or remain unresolved, providing structured evolutionary evidence for subsequent ideation.

\item We introduce \textbf{EvoAgent}, which reasons over the gap-centered trajectories through two complementary signals: \emph{convergent signals} that identify related gaps recurring across multiple trajectories, and \emph{cross-link signals} that connect a method or capability from one trajectory to an unresolved gap in another. These signals enable research ideas grounded in explicit multi-path provenance.

\item Experiments across six AI research topics demonstrate the superior performance of ToI over existing automated ideation methods, particularly in \emph{Novelty} and \emph{Groundedness}. Further ablation studies validate the effectiveness of gap-centered evolution tracing in \textbf{EvoTrace} and cross-trajectory reasoning in \textbf{EvoAgent}.

\end{itemize}

\section{Related Work}
\label{sec:related}

\subsection{Research Idea Generation}

Recent advances in large language models (LLMs) have made automated
research ideation increasingly viable, with several studies showing
that LLM-generated ideas can approach human-level novelty under
expert evaluation~\citep{si-etal-2024-ideabench}. Methodologically,
this progress builds on a broader line of structured LLM reasoning,
which augments chain-of-thought prompting~\citep{wei-etal-2022-cot}
with explicit intermediate structure, such as trees~\citep{yao-etal-2023-tot}
and graphs~\citep{besta-etal-2024-got} of thoughts, or interleaves
reasoning with actions over external
context~\citep{yao-etal-2023-react}. Our work adopts a tree-structured
reasoning process, but the structure is derived from the citation
lineage of the literature rather than from the model's own
deliberation. Existing ideation methods
can broadly be divided according to how they represent and reason
over prior literature.

A first line of work treats literature primarily as retrieved
context, following the broader paradigm of retrieval-augmented and
memory-augmented language models that inject external knowledge at
inference time~\citep{lewis2020retrieval,khandelwal2019generalization,petroni2019language}.
Retrieval-augmented systems collect relevant papers or
summaries and condition generation directly on them.
ResearchAgent~\citep{baek-etal-2024-researchagent} iteratively
refines ideas over literature-grounded knowledge graphs, while
AI-Scientist~\citep{lu-etal-2024-aiscientist} and its successor
AI-Scientist-v2~\citep{yamada-2025-aiscientist-v2} use agentic
pipelines combining reading, summarization, and self-evaluation;
related end-to-end research agents extend this to full experiment
and paper-writing loops~\citep{schmidgall-2025-agentlab,weng-2025-cycleresearcher}.
Other systems focus on hypothesis discovery or trend extrapolation,
including hypothesis-generation frameworks~\citep{qi-etal-2023-hypothesis},
novelty-optimized inspiration and facet-recombination
systems~\citep{wang-etal-2024-scimon,radensky-2024-scideator},
methodology-driven trend forecasting~\citep{hu-etal-2025-nova},
and cross-domain transfer approaches~\citep{garikaparthi-etal-2025-mir}.
Although these methods improve literature awareness, the retrieved
papers are typically exposed as flat context windows or static
knowledge structures, making it difficult to reason explicitly about
how assumptions and gaps evolve across generations of work.

A second line of work attempts to model research evolution more
explicitly. Chain-of-Ideas~\citep{li-etal-2024-chain} organizes
papers into linear evolution chains and generates ideas conditioned
on these trajectories, showing that structural ordering improves
ideation over unordered retrieval. GoAI~\citep{gao-etal-2025-goai}
extends this perspective to citation graphs with typed semantic
relations between papers. A related line encodes the citation graph
into document representations for retrieval and
relatedness~\citep{cohan-2020-specter,ostendorff-2022-scincl},
but uses citations as a similarity signal rather than as a substrate
for reasoning about how contributions evolve. More broadly, methods
that let LLMs reason over graph structure---traversing knowledge
graphs~\citep{sun-etal-2024-tog,luo-etal-2024-rog} or general
graphs~\citep{jin-etal-2024-graphcot}---show that explicit relational
structure can guide multi-step reasoning, though they target
question answering over static graphs rather than ideation over an
evolving research landscape. These approaches move
beyond flat retrieval, but they still largely operate on path-like or
static relational structures. Linear chains compress inherently branching
research landscapes into a single trajectory, while graph relation
labels alone do not capture which assumptions persist, which gaps
remain unresolved, or how solution patterns propagate across
multiple branches.

Our work builds on this line of research but differs in both
representation and reasoning objective. Rather than treating
citations as static relations or forcing evolution into linear
chains, Tree-of-Ideas models literature as a set of branching
evolution trees whose edges explicitly encode how gaps and
mechanisms change across generations. This enables \emph{cross-path
reasoning}: identifying assumptions that independently persist across
multiple branches, or transferring solution mechanisms from one
branch to unresolved gaps on another. In contrast to prior systems,
the central object of reasoning in ToI is therefore not an isolated
paper or a single chain, but the structural interaction between
multiple evolutionary trajectories.

\subsection{Evaluation of Research Ideas}

Evaluating research ideas is inherently challenging because there is
no ground-truth reference and the quality of an idea depends on
multiple dimensions simultaneously. Early work such as
IdeaBench~\citep{si-etal-2024-ideabench} established human-based
evaluation protocols showing that LLM-generated ideas can approach
human experts in novelty while still lagging in feasibility and
practicality. Subsequent benchmarks extended evaluation to broader
settings, including target-paper alignment and downstream execution
outcomes~\citep{wang-etal-2025-ideabench2,chen-etal-2025-mlrbench}.

Two major evaluation paradigms have since emerged. Pairwise
comparison approaches, such as Idea Arena~\citep{li-etal-2024-chain},
perform Elo-style tournaments between generated ideas. These methods
reduce scale calibration issues but provide limited interpretability,
since they cannot decompose performance into dimensions such as
novelty or feasibility. In contrast, multidimensional absolute
scoring offers more interpretable analysis, but suffers from severe
judge-level calibration variance: different LLM judges often apply
systematically shifted scoring scales even when their relative
rankings are similar~\citep{zheng-etal-2023-judging}. Rubric-based
protocols that supply explicit scoring criteria to the judge have
been shown to reduce this variance and improve alignment with human
raters~\citep{liu-2023-geval,kim-2024-prometheus}.

A further complication is that idea-stage evaluation occurs before
experimental validation. \citet{si-etal-2025-ideation-execution}
show that many LLM-generated ideas experience substantial score drops
after implementation, especially on feasibility-related dimensions,
indicating that textual plausibility alone is insufficient for
predicting research value. This creates a systematic tension:
evaluation must reward logically grounded ideas even when empirical
evidence does not yet exist.

Our evaluation protocol addresses these issues through a structured
judgment-style rubric with explicit anchor descriptions for each
score band. In particular, the rubric distinguishes between
unsupported speculation and logically derived claims grounded in
named prior work, allowing idea-stage proposals to receive credit
for rigorous derivational reasoning even before empirical execution.
This reduces ambiguity in judge interpretation and improves scoring
consistency across heterogeneous LLM evaluators.

\section{Methodology}

\label{sec:method}

\subsection{Problem Formulation and the ToI Framework Overview}
  \label{sec:toi_framework}
\paragraph{Problem formulation.}
Given a research topic $q$ and a literature space $\mathcal{L}$, the goal of automated research ideation is to generate a set of structured research ideas $\mathcal{I}$.
In this work, we focus on research opportunities revealed by the evolution of academic literature. As research develops along different trajectories, existing gaps may be inherited, transformed, or remain unresolved, while advances made along one trajectory may provide potential solutions to gaps on another. The objective is therefore to identify such openings across related trajectories and translate them into research ideas grounded in the supporting literature.
\paragraph{The ToI framework overview.}
We propose \textbf{Tree-of-Ideas} (ToI), a three-stage framework that generates research ideas by tracing gap evolution within research trajectories and reasoning across them. As illustrated in Figure~\ref{fig:framework}, the overall process is formulated as
\begin{equation}
(q,\mathcal{L})
\xrightarrow{\text{Construct}}
\mathcal{G}_q
\xrightarrow{\mathrm{EvoTrace}}
\mathcal{T}_r
\xrightarrow{\mathrm{EvoAgent}}
\mathcal{I}.
\end{equation}
  where $\mathcal{G}_q$ is a topic-centered citation graph, $r$ is a frontier paper selected from the graph, and  \(\mathcal{T}_r\) is the corresponding evolution tree containing multiple research trajectories.

ToI constructs \(\mathcal{G}_q\) from literature relevant to \(q\) and identifies frontier papers as entry points for evolution tracing. Given a frontier paper \(r\), \textbf{EvoTrace} reconstructs \(\mathcal{T}_r\) by tracing how research gaps evolve along its citation relations. \textbf{EvoAgent} then reasons across the resulting trajectories, identifying \emph{convergent signals} and \emph{cross-link signals} to generate the final idea set \(\mathcal{I}\). Detailed procedures are presented in the following sections.
  



\subsection{Topic-Centered Citation Graph Construction}
\label{sec:graph-construction}
Given a research topic \(q\) and a literature space \(\mathcal{L}\), 
this stage constructs a directed citation graph $\mathcal{G}_q=(V_q,E_q)$,
where \(V_q\) contains papers relevant to \(q\), and each edge
\((u,v)\in E_q\) indicates that paper \(u\) cites paper \(v\).

The construction begins by retrieving seed papers closely related to
\(q\). Their citation relations are then traced backward to recover the
earlier studies on which they build. During this process, papers that
are insufficiently related to \(q\) are excluded, limiting topic drift
as the citation neighborhood expands. The retained papers and citation
relations are merged into \(\mathcal{G}_q\).

Unlike isolated citation chains, \(\mathcal{G}_q\) preserves the branching relations and shared predecessors required for subsequent evolution tracing. Frontier papers at the recent end of the graph serve as entry points for EvoTrace, which reconstructs local evolution trees and infers how research gaps evolve along their trajectories.

\subsection{EvoTrace: Gap-Centered Evolution Tracing}
\label{sec:evotrace}

Given a frontier paper \(r\) and the topic-centered citation graph
\(\mathcal{G}_q\), EvoTrace constructs an annotated local evolution tree
$\mathcal{T}_r=(V_r,E_r,\phi,\psi)$,
where \(V_r\) and \(E_r\) denote the retained papers and citation
relations, \(\phi\) describes the evolutionary transition associated
with each edge, and \(\psi\) records the research gaps associated with
each paper.

Rather than analyzing papers independently, EvoTrace performs relational
inference over citation-linked paper pairs. Given a predecessor \(v\) and
a citing paper \(u\), it infers how \(u\) advances \(v\), whether the gaps
in \(v\) are addressed, narrowed, inherited, or transformed, and what new
gaps arise. Accordingly, \(\phi(v,u)\) captures the advancement and gap
transition from \(v\) to \(u\), while \(\psi(u)\) records the gaps that
remain or emerge at \(u\). Modeling \(\phi\) at the relation level allows
the same paper to represent different advances over different predecessors.

Starting from (r), EvoTrace recursively follows relevant citations toward earlier work. Although the graph is traversed backward from the frontier paper, each inferred transition is oriented in the evolutionary direction from earlier to later work. Connecting these relation-level transitions produces branching trajectories that trace how advances and research gaps evolve across successive studies. EvoTrace therefore converts a local citation structure into an explicit gap-evolution structure, rather than a collection of independent paper-level summaries.

The resulting \(\mathcal{T}_r\) provides the structured trajectories used
by EvoAgent to identify convergent and cross-link signals across different
branches. EvoTrace focuses on modeling evolution within individual
trajectories, while cross-trajectory opportunity discovery is performed by
EvoAgent. The corresponding relational inference prompts are provided in
Appendix~\ref{app:prompts:evotrace_pair}.

\subsection{EvoAgent: Cross-Trajectory Idea Generation}
    \label{sec:evoagent}
    
Given a research topic \(q\) and the gap-centered evolution tree
\(\mathcal{T}_r\) produced by EvoTrace, EvoAgent generates a set of
research ideas \(\mathcal{I}_r\) through cross-trajectory reasoning.
While EvoTrace models how advances and research gaps evolve within
individual trajectories, EvoAgent compares these trajectories to
identify research opportunities that cannot be exposed by any single lineage alone.
\paragraph{Trajectory Representation.}
EvoAgent extracts the retained evolutionary paths
\(\mathcal{P}_r\) from \(\mathcal{T}_r\). Each path is summarized by
its sequence of evolutionary transitions \(\phi\), the unresolved or
emerging gaps recorded by \(\psi\), and the papers supporting these
transitions. This representation preserves the gap-evolution structure
inferred by EvoTrace while making different trajectories directly
comparable.
\paragraph{Cross-Trajectory Signal Discovery.}
EvoAgent compares the trajectories to identify two classes of
signals. \emph{Convergent signals} capture structural patterns shared
across trajectories and include three forms:
(1) \emph{shared gaps}, where the same or related gaps persist across
multiple trajectories;
(2) \emph{common actions}, where different trajectories repeatedly
adopt the same research action or solution pattern, revealing a shared
design choice or underlying assumption; and
(3) \emph{frontier races}, where distinct approaches pursue the same
frontier objective in parallel.
These patterns represent convergence at the problem, action, and
objective levels, respectively.
\emph{Cross-link signals} take the form of
\emph{capability--gap bridges}, where a capability developed along one
trajectory can potentially address an unresolved gap exposed by another.

Each detected signal is converted into a candidate research direction
together with its supporting trajectories and papers. A direction is
retained only when its evidence spans at least two distinct trajectories;
otherwise, it is treated as a single-lineage continuation rather than a
cross-trajectory opportunity.
\paragraph{Idea Instantiation and Validation.}
For each retained direction, EvoAgent generates a structured research
idea consisting of a candidate contribution, a method sketch, and its
trajectory-level provenance. The candidate is refined through an
internal review and checked for novelty against the broader literature
beyond \(\mathcal{T}_r\). Candidates that fail validation are discarded.
Finally, a diversity filter removes near-duplicate ideas according to
method-level embedding similarity, yielding the final idea set
\(\mathcal{I}_r\).


\begin{algorithm}[t]
  \small
  \caption{EvoAgent: Cross-Trajectory Ideation}
  \label{alg:evoagent}
  \begin{algorithmic}[1]
  \REQUIRE Gap-centered evolution tree
  $\mathcal{T}_r=(V_r,E_r,\phi,\psi)$, research topic $q$
  \ENSURE Research ideas $\mathcal{I}_r$

  \STATE Extract evolutionary trajectories
  $\mathcal{P}_r$ from $\mathcal{T}_r$.
  \IF{$|\mathcal{P}_r|<2$}
      \RETURN $\emptyset$
  \ENDIF

  \FOR{each trajectory $p\in\mathcal{P}_r$}
      \STATE Construct a topic-conditioned summary $\mathcal{S}_p$
      from its transitions, gaps, and supporting papers.
  \ENDFOR

  \STATE Initialize candidate directions
  $\mathcal{D}\leftarrow\emptyset$.

  \STATE Detect \textit{Convergent signals} from
  $\{\mathcal{S}_p\}_{p\in\mathcal{P}_r}$.
  \STATE Detect \textit{cross-link signals} across
  distinct trajectories.
  \STATE Convert each valid signal into a direction $d$ with
  its supporting trajectory set $\operatorname{Prov}(d)$.
  \STATE Retain only directions satisfying
  $|\operatorname{Prov}(d)|\geq2$.

  \STATE Initialize $\mathcal{I}_r\leftarrow\emptyset$.
  \FOR{each direction $d\in\mathcal{D}$}
      \STATE Generate an idea $i$ conditioned on $q$, $d$, and
      its supporting trajectory evidence.
      \STATE Refine $i$ through internal review.
      \STATE Check the novelty of $i$ against the broader literature.
      \IF{$i$ passes review and novelty checking}
          \STATE Add $(i,\operatorname{Prov}(d))$ to $\mathcal{I}_r$.
      \ENDIF
  \ENDFOR

  \STATE Remove near-duplicate ideas in $\mathcal{I}_r$ using
  method-level embedding similarity.
  \RETURN $\mathcal{I}_r$
  \end{algorithmic}
\end{algorithm}

\section{Experiments}
\label{sec:experiments}

\subsection{Experimental Setup}
\label{sec:setup}

\paragraph{Implementation details.}
All automatic methods, including ToI, use DeepSeek-V4-Flash as the backbone model~\citep{deepseek2026v4flash}. We evaluate each method on six research topics (see Appendix~\ref{sec:appendix-topics}), with five ideas generated per topic. As described in Section~\ref{sec:graph-construction}, the topic-specific literature space is constructed from the Semantic Scholar API~\citep{ammar-etal-2018-construction} and a pre-built citation graph of papers from major AI venues (NeurIPS, ICML, ACL, EMNLP, ICLR, CVPR, AAAI, etc.). Starting from topic-relevant frontier papers, we recursively expand their references to a maximum depth of three. All literature-based methods access the same topic-specific literature space, whereas Direct Prompting receives only the topic description. We additionally include five Real Paper ideas per topic as published-idea references.



\paragraph{Baselines.}
  \label{sec:baselines}
  We compare ToI against five automatic ideation methods and one
  published-idea reference:

  \begin{itemize}[nosep,leftmargin=*]
    \item \textbf{Direct Prompting}: LLM generates ideas from the
      topic description alone, without literature context.
    \item \textbf{RAG}: Standard retrieval-augmented
      generation~\citep{lewis2020retrieval,yan2024corrective}---paper
      abstracts are retrieved and concatenated as context for idea
      generation.
    \item \textbf{CoI-Agent} \citep{li-etal-2024-chain}: Constructs
      linear research-trend chains from anchor papers, then generates
      ideas conditioned on chain context.
    \item \textbf{AI-Scientist} \citep{lu-etal-2024-aiscientist}: A
      multi-step agentic framework that reads, summarizes, and generates
      ideas with self-evaluation.
    \item \textbf{ResearchAgent} \citep{baek-etal-2024-researchagent}:
      Iteratively generates and refines research ideas over scientific
      literature using knowledge graphs.
    \item \textbf{Real Paper}$^\dagger$: Core ideas extracted from
      representative published papers, serving as a human-level
      reference rather than a competing system.
  \end{itemize}

\subsection{Evaluation Setup}
  \label{sec:eval-setup}

  \paragraph{Evaluation dimensions.}
  We evaluate each generated idea on five dimensions---Novelty,
  Significance, Groundedness, Feasibility, and Expected
  Effectiveness---using a 10-point scale (definitions in
  Table~\ref{tab:eval-dims}, full per-band rubrics in
  Table~\ref{tab:rubrics}).

  \paragraph{Model-based evaluation.}
  As idea generation lacks ground-truth references, rendering automatic
  metrics inapplicable, we adopt an ensemble of three frontier LLMs as
  judges: GPT-5.5~\citep{openai2026gpt55}, Claude
  Opus~4.7~\citep{anthropic2026claudeopus47}, and
  Grok~4.3~\citep{xai2026grok43}.
  Each judge independently scores every idea across the five dimensions
  above (prompt details in Appendix~\ref{sec:appendix-eval-prompt}).

\paragraph{Human evaluation.}
We invite five researchers with backgrounds in AI to score a subset
of generated ideas along the same five dimensions using a 10-point
scale. Reviewers independently evaluate anonymized outputs presented
in random order.

\subsection{Experimental Results}  
\label{sec:main-results}
  \paragraph{Main Results}

  Table~\ref{tab:main-results} reports per-dimension scores averaged over three LLM judges on a 10-point scale. ToI achieves the best overall performance among all automatic methods, with an average score of $6.27$, compared with $5.36$ for the strongest baseline, ResearchAgent. It ranks first among automatic methods across all five dimensions, achieving $6.36$ in \emph{Novelty}, $5.71$ in \emph{Significance}, $7.00$ in \emph{Groundedness}, $6.59$ in \emph{Feasibility}, and $5.68$ in \emph{Effectiveness}. The particularly strong results in Novelty and \emph{Groundedness} support our central hypothesis that reconstructing scholarly evolution and reasoning across multiple trajectories enables the discovery of research ideas that are both less obvious and more firmly connected to prior work. Its consistent performance across the remaining dimensions further indicates that these gains do not come at the expense of overall proposal quality. ToI also reaches an aggregate score comparable to the Real Paper reference ($6.29$). In contrast, Direct Prompting obtains a substantially lower \emph{Groundedness} score ($3.83$), highlighting the importance of literature-grounded context for credible automated research ideation.

  
  \begin{table}[t]
      \centering
      \renewcommand{\arraystretch}{1.15}
      \resizebox{\columnwidth}{!}
      {%
      \begin{tabular}{@{}lcccccc@{}}
        \toprule
        \textbf{Method} & \textbf{Nov.} & \textbf{Sig.} &
  \textbf{Gnd.}
        & \textbf{Fea.} & \textbf{Eff.} & \textbf{Avg.} \\
      \midrule
        Real Paper$^\dagger$ & 5.96 & 6.46 & 6.01 & 6.60 & 6.43 &
  6.29 \\
        \midrule
        Direct Prompting & 4.59 & 5.19 & 3.83 & 5.20 & 4.71 & 4.70
   \\
        RAG              & \underline{5.08} & 5.07 &
  \underline{5.70} & 5.57 & 4.63 & 5.21 \\
        CoI-Agent~\citep{li-etal-2024-chain}        & 4.84 & 5.11
  &
  5.59 & 5.83 & 4.89 & 5.25 \\
        AI-Scientist~\citep{lu-etal-2024-aiscientist}     & 5.07 &
  5.02 & 5.30 & \underline{6.45} & 4.81 & 5.33 \\
        ResearchAgent~\citep{baek-etal-2024-researchagent}    &
  4.71
  & \underline{5.54} & 5.42 & 5.87 & \underline{5.27}
  &
  \underline{5.36} \\
        \midrule
        \textbf{ToI (Ours)} & \textbf{6.36} & \textbf{5.71} &
  \textbf{7.00} & \textbf{6.59} & \textbf{5.68} & \textbf{6.27}
   \\
        \bottomrule
      \end{tabular}%
      }
      \caption{Model-based evaluation: scores averaged across
  three LLM judges (GPT-5.5, Claude Opus~4.7, Grok~4.3) over 6
  topics
      (30 ideas per method).
      \textbf{Bold} marks the best automatic method per column;
      \underline{underline} marks the second.
      $^\dagger$Real Paper is a human reference, not a competing
  system.
      Scores are on a 10-point scale.}
      \label{tab:main-results}
  \end{table}

  \paragraph{Main Results (Human)}
  \label{sec:human-eval}
  Table~\ref{tab:human-eval} reports human ratings on the same five dimensions and 10-point scale, based on evaluations from five domain experts across six topics. CoI-Agent and ToI achieve the strongest overall performance among the automatic methods, with average scores of $6.71$ and $6.62$, respectively, while the remaining baselines range from $5.68$ to $6.08$. ToI ranks first in \emph{Novelty} ($6.78$) and \emph{Groundedness} ($7.17$), and ties for first in Significance ($6.39$). These results are consistent with the intended roles of EvoTrace and EvoAgent: tracing unresolved gaps across scholarly evolution supports the discovery of less obvious research directions, while citation-anchored evolutionary trajectories provide stronger grounding in prior work. ToI obtains a lower \emph{Feasibility} score ($6.28$), suggesting that ideas derived from cross-trajectory synthesis may be more ambitious and therefore more difficult to operationalize immediately. At the aggregate level, ToI performs comparably to the Real Paper reference ($6.55$), although the reference remains stronger in \emph{Feasibility} and \emph{Effectiveness}. This pattern is consistent with evaluating proposals at the idea stage, before implementation and empirical validation.

\subsection{Ablation study}
\label{sec:analysis-section}
We conduct ablation studies to examine three key components of ToI:
the multi-path evolutionary structure, gap modeling in EvoTrace, and
cross-trajectory signal discovery in EvoAgent. All variants use the same
backbone model, literature space, generation setting, and evaluation
protocol as the full framework.

\begin{table}[!]
  \centering
  \renewcommand{\arraystretch}{1.15}
  \resizebox{\columnwidth}{!}{%
  \begin{tabular}{@{}lcccccc}
    \toprule
    \textbf{Method} & \textbf{Nov.} & \textbf{Sig.} &
    \textbf{Gnd.} & \textbf{Fea.} & \textbf{Eff.}
    & \textbf{Avg.} \\
    \midrule
    Real Paper$^\dagger$ & 5.95 & 6.44& 6.83 & 6.72& 6.83& 6.55 \\
    \midrule
    Direct Prompting & 5.44 & 5.89 & 5.22 & 5.95 & 5.89 & 5.68 \\
    RAG              & 6.11 & 5.95 & 6.05 & 5.95 & 5.94 & 6.00 \\
    AI-Scientist~\citep{lu-etal-2024-aiscientist}     
                     & 5.94 & 5.89 & 5.78 & \textbf{6.56}& 6.22 & 6.08 \\
    ResearchAgent~\citep{baek-etal-2024-researchagent} 
                     & 6.05 & 6.11& 6.17 & 6.22 & 5.72 & 6.06 \\
    CoI-Agent~\citep{li-etal-2024-chain}        
                     & \underline{6.55} & \underline{6.39}& \underline{7.05} & \underline{6.55}& \textbf{7.00} & \textbf{6.71} \\
    \midrule
    \textbf{ToI (Ours)} 
                     & \textbf{6.78} & \textbf{6.39}& \textbf{7.17} & 6.28 & \underline{6.50}& \underline{6.62} \\
    \bottomrule
  \end{tabular}%
  }
  \caption{Human-based evaluation: 5 experts across 6 topics
  (30 ideas per method).
  \textbf{Bold} marks the best automatic method per column;
  \underline{underline} marks the second.
  $^\dagger$Real Paper is a human reference, not a competing system.
  Scores are on a 10-point scale.}
  \label{tab:human-eval}
\end{table}

\paragraph{Multi-Path Structure as a Reasoning Inductive Bias}
To isolate the effect of evolutionary structure, we compare the full ToI
with two degraded variants. \textit{Chain-only} retains only the longest
depth-first path from each frontier paper, while \textit{frontier-only}
removes the evolutionary paths and uses only frontier papers. As shown in
Table~\ref{tab:ablation}, performance consistently decreases as structural
context is removed: full ToI (\(6.04\)) outperforms chain-only (\(5.85\))
and frontier-only (\(4.77\)). The largest reductions occur in Novelty and
Groundedness, indicating that multiple evolutionary paths expose both more
diverse research opportunities and stronger historical support. Chain-only
achieves the highest Feasibility, suggesting that a narrower context tends
to produce more conservative and readily executable proposals.

\paragraph{Gap Modeling in EvoTrace.}
We examine whether explicitly tracing research gaps contributes beyond
recovering paper-level evolutionary relations. The \textit{w/o Gap} variant
retains the same evolution-tree structure but
removes the gap annotations inferred along each trajectory. Compared with the ToI, reduces the average score from 6.27 to 5.55, with consistent degradation across all five evaluation dimensions, which indicates that recovering evolutionary relations alone is insufficient for effective ideation. Explicit gap modeling helps identify unresolved problems along research trajectories and provides concrete targets for transforming evolutionary evidence into research ideas.

\paragraph{Cross-Trajectory Signals in EvoAgent.}
We ablate the signal discovery mechanism used by EvoAgent.
The \textit{w/o Signals} variant receives the same trajectory summaries
but directly generates ideas without detecting explicit cross-trajectory
patterns. \textit{Convergent-only} retains shared gaps, common actions,
and frontier races, while removing capability--gap bridges.
Conversely, \textit{Cross-link-only} retains only capability--gap bridges
across trajectories.
The results show that removing explicit signal discovery decreases the average score from $6.27$ to $5.62$. Retaining either signal type partially recovers performance, with \textit{Convergent-only} and \textit{Cross-link-only} reaching $5.71$ and $5.81$, respectively, but both remain below the full model. This demonstrates that explicit cross-trajectory signals provide more informative guidance than directly generating from trajectory summaries, while convergent and cross-link signals capture complementary opportunities that are most effective when used jointly.


\begin{table}[t]
  \centering
  \renewcommand{\arraystretch}{1.12}
  \resizebox{\columnwidth}{!}{%
    \begin{tabular}{@{}lccccc|c@{}}
    \toprule
    \textbf{Variant}
    & \textbf{Nov.}
    & \textbf{Sig.}
    & \textbf{Gnd.}
    & \textbf{Fea.}
    & \textbf{Eff.}
    & \textbf{Avg.} \\
    \midrule

    ToI (full)
    & 6.36 & 5.71 &
  7.00 & 6.59 & 5.68 & 6.27 \\\hline
 Chain-only
    & 5.93 & 5.57 & 5.65 & 6.93 & 5.38 & 5.85 \\
Frontier-only
    & 5.05 & 4.75 & 3.89 & 6.01 & 4.27 & 4.77 \\
 w/o Gap
    & 5.80 & 5.20 & 6.20 & 5.80 & 4.60 & 5.55 \\
 w/o Signals
    & 5.11 & 4.89 & 6.67 & 7.00 & 4.89 & 5.62 \\
Convergent-only
    & 6.00 & 5.17 & 6.54 & 5.94 & 4.77 & 5.71 \\
 Cross-link-only
    & 5.57 & 5.29 & 7.00 & 6.14 & 5.14 & 5.81 \\
    \bottomrule
  \end{tabular}%
  }
  \caption{
  Ablation study of the evolutionary structure, gap modeling, and
  cross-trajectory signals in ToI. Results are averaged over six research
  topics with 30 ideas per configuration and reported on a 10-point scale.
  }
  \label{tab:ablation}
\end{table}








\subsection{Case study: cross-path idea on Diffusion.}

\label{sec:case-study}

Table~\ref{tab:case-study} traces the full ToI pipeline on the \textbf{Diffusion Models} topic.
We illustrate how ToI derives an idea from evidence distributed across
multiple research trajectories. Starting from \emph{Simple and Critical
Iterative Denoising} (2025), EvoTrace reconstructs two mainline branches:
one develops Critic-guided denoising, while the other introduces
score-based corrector steps. Although both improve the reverse process,
they retain the fixed mask-absorbing forward schedule inherited from
earlier discrete diffusion models. EvoAgent identifies this recurring,
untouched assumption as a convergent gap and proposes \textbf{NF-Diff},
which learns state-conditioned, per-element forward transition
distributions. Considering both trajectories strengthens the evidence
that the forward process is a shared structural bottleneck rather than
a local omission along one branch.


\begin{table}[t]
\centering
\small
\renewcommand{\arraystretch}{1.3}
\begin{tabular}{@{}p{2.1cm}p{5.2cm}@{}}
\toprule
\textbf{Stage} & \textbf{Content} \\
\midrule
\textbf{Frontier paper} &
\emph{Simple and Critical Iterative Denoising} (2025): recasts discrete graph diffusion as element-wise Critic-guided denoising. \\
\midrule
\textbf{EvoTrace} \newline \textbf{(Trajectory evidence)} &
\textbf{Branch 1} — \emph{Think While You Generate} (2024): replaces planner--denoiser with a single Critic network. \newline
\textbf{Branch 2} — \emph{Informed Correctors} (2024): adds score-based corrector steps to iterative denoising. \newline
Both branches improve the reverse process; neither questions the forward noise schedule. \\
\midrule
\textbf{Convergent gap} \newline \texttt{priority=0.95} &
Every predecessor and the frontier assume a fixed, hand-crafted mask-absorbing noise schedule applied uniformly across all elements. No branch learns transition dynamics from data. \\
\midrule
\textbf{EvoAgent} \newline \textbf{direction} &
\emph{Learn the forward corruption trajectory from data, replacing the fixed noise schedule.} \\
\midrule
\textbf{Generated idea} &
\textbf{NF-Diff} (Neural Forward Diffusion): parameterizes the forward transition kernel via a small hypernetwork conditioned on the current noisy graph, outputting per-element distributions over the discrete vocabulary. The reverse process inverts this learned, data-adaptive corruption jointly. \\
\bottomrule
\end{tabular}
\caption{ToI pipeline trace on the Diffusion topic. The convergent gap (row 3) is only visible when both mainline branches are held simultaneously; a single-chain method following either branch alone would not identify the fixed forward process as the shared structural bottleneck.}
\label{tab:case-study}
\end{table}

\section{Conclusion}
\label{sec:conclusion}
We introduced \textbf{Tree-of-Ideas} (ToI), which frames automated scientific ideation as cross-trajectory reasoning over scholarly evolution. By reconstructing evolution trees and tracing gaps across multiple research trajectories, ToI uses citation structure as explicit reasoning context rather than flat retrieval evidence. Experiments across six AI topics show that ToI achieves the strongest overall performance among automatic methods, particularly in Novelty and Groundedness, while ablations confirm the value of multi-path evolutionary structure. These results demonstrate the potential of scholarly evolution as a structured basis for generating grounded and original research ideas.

\section*{Limitations}
Our evaluation is conducted at the idea stage without downstream implementation or empirical validation; therefore, highly rated ideas may not necessarily lead to successful research outcomes. In addition, ToI relies on the completeness and quality of the constructed citation graph, and missing or weakly connected trajectories may limit the evolutionary evidence available for cross-trajectory reasoning.


\clearpage
\bibliography{custom}

\clearpage
\appendix
  \section{EvoGraph Construction Details}
  \label{sec:appendix-evograph}

  The citation graph $\mathcal{G}_q = (V_q, E_q)$ is constructed as
  follows. Each node $v \in V_q$ stores lightweight bibliographic
  metadata: title, abstract, publication year, and an open-access PDF
  URL where available. Full PDF content is not retrieved at this stage;
  it is fetched on demand by EvoTrace during gap analysis. Each
  directed edge $(u, v) \in E_q$ indicates that paper $u$ cites paper
  $v$, so edges point from a paper toward its cited predecessors. No
  semantic evolution labels are assigned during construction.

  Seed papers are retrieved from the Semantic Scholar
  API~\citep{ammar-etal-2018-construction} and a pre-built venue index
  covering accepted papers from top AI venues (NeurIPS, ICML, ACL,
  EMNLP, ICLR, CVPR, AAAI, etc.), restricted to papers published after
  May 2025. $\mathcal{G}_q$ is then grown by a bounded $K{=}3$-hop
  BFS backward traversal: at each hop, leaf nodes are ranked by the
  number of topic-labeled papers that directly cite them, and the top
  30 are expanded; each expanded node retains at most 30 references.
  If the Semantic Scholar API returns no references for a seed paper, a
  GROBID-based PDF fallback is attempted: the open-access PDF is
  downloaded and its reference list is parsed and resolved back to
  Semantic Scholar IDs. A paper is admitted to the confirmed node set
  only if at least one outgoing citation edge is successfully
  established; otherwise it is rejected from the seed set. Survey and
  review papers are filtered throughout by keyword matching on title
  and abstract.

  \section{Frontier Selection Details}
  \label{app:frontier}

  The frontier paper $r$ that anchors EvoTrace
  (\S\ref{sec:evotrace}) is selected from $\mathcal{G}_q$ by a
  three-step procedure: candidate filtering, structural feature
  extraction, and LLM-based routing.

  \paragraph{Candidate filtering.}
  We restrict candidates to papers tagged with topic $q$ in the most
  recent two publication years available in $\mathcal{G}_q$. Survey
  papers are removed, as are papers without at least one older cited
  predecessor inside $\mathcal{G}_q$ (such papers admit no non-trivial
  ancestral subgraph and thus no evolution trajectory to reason about).
  If this filter returns an empty set, we fall back to the first
  twenty topic-tagged papers as candidates.

  \paragraph{Trajectory features.}
  For each surviving candidate $c$, we trace its ancestral subgraph
  backwards through citation edges (BFS depth $\leq 5$, year cutoff
  $\geq 2022$, at most $8$ predecessors per node, retaining the most
  recent ones) and compute four structural features:
  \begin{itemize}[leftmargin=*,topsep=2pt,itemsep=1pt]
    \item \textbf{depth}: length of the longest predecessor chain
          rooted at $c$;
    \item \textbf{reach}: total number of ancestors reachable from
          $c$ within the cutoffs;
    \item \textbf{branching}: number of nodes in the subgraph with
          multiple predecessors (a proxy for methodological
          confluence, and the structural signature that motivates
          the DAG-to-tree unfolding in \S\ref{sec:evotrace});
    \item \textbf{span}: publication-year range of the subgraph.
  \end{itemize}

  \paragraph{LLM router.}
  Candidates are split into batches of size $8$. Each batch is sent
  to an LLM router in parallel
  (Appendix~\ref{app:prompts:frontier}) together with the candidate
  title, year, structural features, and abstract; the router returns
  $\lceil k/2 \rceil$ winners per batch. Survivors from all batches
  are deduplicated and the top-$k$ are kept as frontier papers
  (default $k=5$). When the router fails or returns too few items,
  we back off to a heuristic score
  $0.4 \cdot \mathrm{depth} + 0.1 \cdot \mathrm{reach}
   + 0.5 \cdot \mathrm{branching}$
  that emphasizes deep, structurally branching trajectories.

  \section{Prompt Templates}
  \label{app:prompts}

  This section documents the prompt templates used in the ToI
  pipeline. We treat these prompts as one concrete instantiation of
  the structured-inference oracles defined in
  \S\ref{sec:evotrace}; the algorithmic contribution lies in the
  inference structure (the trajectory features, the edge-level
  $\phi$ / node-level $\psi$ decomposition, the cross-branch
  evidence rule), not in any specific wording. Templates below are
  shown in lightly abridged form for space; the full templates are
  released with our code.

  \subsection{Frontier Routing Prompt}
  \label{app:prompts:frontier}

  Used by the LLM router in Appendix~\ref{app:frontier} to select
  the top-$k$ frontier papers within each candidate batch. The
  router sees the candidate's title, year, four structural
  trajectory features (depth, reach, branching, span), and abstract.
  Source: \texttt{prompts/evotrace\_agent\_prompts.py}.

  \begin{promptbox}[Frontier Routing]
  You are a research strategist selecting the most promising research
  trajectories for novel idea generation.

  \medskip
  \textbf{Research Topic:} \texttt{\{topic\}}

  \medskip
  \textbf{[Candidates]}\quad each candidate listed with its
  graph-structural features:
  \begin{itemize}[nosep,leftmargin=*]
    \item \texttt{depth}: how many generations of predecessor papers
          can be traced back (deeper = richer evolution story);
    \item \texttt{reach}: total papers reachable by tracing the
          citation chain backwards (more = broader context);
    \item \texttt{branch}: number of convergence points where multiple
          research threads meet;
    \item \texttt{span}: years of research history covered by the
          trajectory;
    \item \texttt{relevance}: semantic similarity to the topic ($0$--$1$).
  \end{itemize}
  \quad \texttt{\{candidates\_str\}}

  \medskip
  \textit{Task:} Select exactly $k$ papers that produce the most
  diverse and insightful evolution trees for novel idea generation.

  \medskip
  \textit{Selection criteria:}
  \begin{enumerate}[nosep,leftmargin=*]
    \item DIVERSITY: chosen papers should represent different research
          sub-directions (avoid selecting papers that study the same
          specific problem).
    \item DEPTH: prefer papers with deeper predecessor chains.
    \item NOVELTY POTENTIAL: trajectories likely to reveal unresolved
          gaps and open questions.
    \item RELEVANCE: must be meaningfully connected to the research topic.
  \end{enumerate}

  \medskip
  Return ONLY a JSON object:
\begin{Verbatim}[fontsize=\scriptsize,breaklines=true,breakanywhere=true]
{"selected": [list of integer indices],
 "reasoning": "..."}
\end{Verbatim}
  \end{promptbox}

  \subsection{EvoTrace Per-Pair Evolution Prompt}
  \label{app:prompts:evotrace_pair}

  Used to instantiate the edge annotation $\phi$ and the residual
  gap label $\psi$ for a single (child, parent) edge in
  $\mathcal{T}_r$ (\S\ref{sec:evotrace}). For each edge, the LLM
  produces (i) an \emph{evolution trace} describing the
  mechanism-level contrast, and (ii) a \emph{gap analysis}
  identifying what the child paper did not close from the parent.
  Source: \texttt{prompts/evotrace\_v2\_prompts.py}.

  \begin{promptbox}[EvoTrace Per-Pair Evolution]
  You are analyzing how paper $A$ evolved from one of its predecessors
  $B$, on the topic \texttt{\{topic\}}.

  \medskip
  Two outputs are required:
  \begin{itemize}[nosep,leftmargin=*]
    \item \texttt{evolution\_trace} --- the CONTRAST: how $A$ evolved from $B$.
    \item \texttt{gap\_analysis} --- the MAIN: what $A$ did NOT close from $B$.
  \end{itemize}

  \medskip
  \textbf{[Paper $A$ --- current]}\quad title, contribution, method,
  self-stated limitation. \\
  \textbf{[Paper $B$ --- predecessor]}\quad title, contribution, method,
  self-stated limitation.

  \medskip
  \textit{Plan (think through these before producing JSON):}
  \begin{enumerate}[nosep,leftmargin=*]
    \item Identify $B$'s specific problem (name the mechanism, not
          ``$B$ has issues'').
    \item Identify how $A$ specifically addresses it (name $A$'s
          mechanism, not its outcome).
    \item Identify what $A$ did NOT change from $B$ (which assumption
          $A$ silently inherits).
    \item Identify what $A$ failed to close from $B$ (which of $B$'s
          limitations is still open).
    \item Categorize the gap as one of \{\texttt{assumption\_inherited},
          \texttt{scope\_unclosed}, \texttt{structural\_unestablished},
          \texttt{self\_admitted}\}.
    \item Categorize the evolution direction as one of
          \{\texttt{structural}, \texttt{scaling}, \texttt{objective},
          \texttt{data}, \texttt{reasoning\_pattern},
          \texttt{verification}\}.
  \end{enumerate}

  \medskip
  Output ONLY valid JSON:
\begin{Verbatim}[fontsize=\scriptsize,breaklines=true,breakanywhere=true]
{
  "evolution_trace": {
    "parent_problem":      "...",
    "child_solution":      "...",
    "mechanism_delta":     "...",
    "evolution_direction": "..."
  },
  "gap_analysis": {
    "remaining_gap": "...",
    "what_persists": "...",
    "gap_type":      "..."
  }
}
\end{Verbatim}
  \end{promptbox}

  The two output blocks correspond directly to the formal
  annotations defined in \S\ref{sec:evotrace}: the
  \texttt{evolution\_trace} block populates the \emph{direction}
  axis of $\phi$ (a six-way categorical) and its mechanism delta;
  the \texttt{gap\_analysis} block populates the \emph{gap status transition}
  axis of $\phi$ (the four-way \texttt{gap\_type}) together with
  the residual gap that contributes to $\psi$ at the parent node.

  \subsection{Phase 2: Forest Signal Aggregation}
  \label{app:prompts:phase2}

  Used after EvoTrace populates $\phi$ and $\psi$ across the forest
  of evolution trees. This single LLM call consumes the per-tree
  signal pack (residual gaps, shared cross-branch gaps, recent
  edge actions, action distributions, frontier metadata) and emits
  a ranked list of research \emph{directions}, each citing
  $\geq\!2$ pieces of evidence at either the cross-branch or
  cross-tree level. Source:
  \texttt{prompts/evo\_agent\_prompts.py}.

  \begin{promptbox}[Phase 2: Forest Signal Aggregation]
  You are scanning a FOREST of paper evolution trees to find FUTURE
  research directions.

  \medskip
  \textbf{Topic area:} \texttt{\{topic\}} \\
  \textbf{Forest size:} \texttt{\{n\_trees\}} independent paper-trees.

  \medskip
  \textbf{[Per-tree signal pack]}\quad \texttt{\{forest\_text\}}
  --- residual gaps, shared cross-branch gaps, recent edge actions,
  action distribution, \texttt{frontier\_solves},
  \texttt{competing\_frontiers}, composite-signal flags.

  \medskip
  Convergent patterns can manifest at TWO levels:
  \begin{itemize}[nosep,leftmargin=*]
    \item \textbf{cross-TREE}: the same pattern recurs across $\geq 2$
          trees (only when \texttt{n\_trees} $\geq 2$).
    \item \textbf{cross-BRANCH}: within ONE tree, the same pattern
          recurs across $\geq 2$ evolution branches.
  \end{itemize}
  Both levels are first-class evidence.

  \medskip
  \textit{Run FOUR signal scans, in order:}
  \begin{enumerate}[nosep,leftmargin=*]
    \item \texttt{convergent\_gap} --- a residual gap recurring across
          $\geq 2$ evidence units.
    \item \texttt{convergent\_action} --- $\geq 2$ evidence units'
          recent actions point in the same direction (the field's
          collective trajectory; the next bottleneck is the implied
          problem after this trajectory completes).
    \item \texttt{bridge} --- one unit's \texttt{frontier\_solves}
          matches another unit's \texttt{residual\_gap} (cross-tree
          OR cross-branch).
    \item \texttt{frontier\_race} --- $\geq 2$ frontiers are recent
          ($\sim$2 yr), small \texttt{depth\_to\_oldest}, and list
          \texttt{competing\_frontiers}.
  \end{enumerate}

  \medskip
  \textit{Absolute rules:}
  \begin{itemize}[nosep,leftmargin=*]
    \item Each direction MUST cite $\geq 2$ evidence units. Use
          \texttt{evidence\_branch\_ids} (intra-tree) OR
          \texttt{evidence\_tree\_ids} (cross-tree). At least ONE
          array must contain $\geq 2$ entries. EXCEPTION:
          \texttt{frontier\_race} may degrade to a single unit (low
          priority).
    \item NON-OBVIOUSNESS GATE: reject any direction whose mechanism
          reduces to a textbook composition of two well-known methods.
          Name the prior method the obvious composition collapses
          onto, and name the structural property your direction
          targets that the composition cannot deliver.
    \item Reject directions whose contribution is a benchmark,
          taxonomy, dataset release, or empirical study; the output
          must be a METHOD paper.
  \end{itemize}

  \medskip
  Output ONLY valid JSON:
\begin{Verbatim}[fontsize=\scriptsize,breaklines=true,breakanywhere=true]
{
  "directions": [{
    "direction_id":         "d_1",
    "type":                 "convergent_gap |
                             convergent_action |
                             bridge | frontier_race",
    "priority":             0.0,
    "title":                "...",
    "rationale":            "...",
    "evidence_tree_ids":    ["..."],
    "evidence_branch_ids":  ["..."],
    "non_obvious_property": "...",
    /* type-specific fields:
       convergent_gap    -> meta_gap,
                            shared_mechanism
       convergent_action -> trend, trend_endpoint,
                            implied_next_problem
       bridge            -> source_*, target_*,
                            what_to_transfer
       frontier_race     -> active_battle,
                            common_failure_mode
    */
  }]
}
\end{Verbatim}

  Aim for 3--8 directions, sorted by priority.
  \end{promptbox}

  The cross-branch evidence rule
  (\texttt{evidence\_branch\_ids} with $\geq\!2$ entries) is what
  allows the single-tree configuration ($k\!=\!1$) to act as the
  default mode in our experiments: a single $\mathcal{T}_r$ already
  exposes multiple evolution branches, and convergence across them
  is sufficient evidence even when the forest contains only one
  tree.

  \subsection{EvoAgent Ideation Prompt}
  \label{app:prompts:evoagent}

  Used in the final stage $F_{\mathrm{EvoAgent}}$ to convert one
  ranked direction (plus the supporting idea cards retrieved from
  $\mathcal{T}_r$) into a concrete research idea. Source:
  \texttt{prompts/evo\_agent\_prompts.py}.

  \begin{promptbox}[EvoAgent Ideation]
  You are a research idea generation expert. Based on the following
  research direction prediction and related paper information,
  generate a research idea.

  \medskip
  \textbf{Topic:} \texttt{\{topic\}}

  \medskip
  \textbf{[Research Direction]}\quad direction, rationale, builds-on,
  persistent gap (the gap that survived multiple evolution steps in
  this tree), why-still-open (the shared assumption blocking it).

  \medskip
  \textbf{[Forest-level motivation --- when present]}\quad direction
  type, meta-gap, shared mechanism, implied next problem, what to
  transfer, active battle / failure mode.

  \medskip
  \textbf{[Related Papers --- Idea Cards]}\quad
  \texttt{\{cards\_text\}}.

  \medskip
  \textbf{[Latest Related Work]}\quad (optional, off by default) ---
  must NOT duplicate at the mechanism level.

  \medskip
  Respond with a JSON object following this schema:
\begin{Verbatim}[fontsize=\scriptsize,breaklines=true,breakanywhere=true]
{
  "title":        "...",
  "motivation":   "<concrete limitation of
                   existing methods + root
                   cause; cite specific Idea
                   Cards>",
  "key_insight":  "<one sentence: structural
                   property / invariant /
                   theoretical grounding that
                   makes the approach work>",
  "method":       "<connected prose, 4-7
                   sentences. Lead with what
                   the method DOES at mechanism
                   level (inputs/outputs +
                   structural property
                   exploited). Hyperparameters
                   appear next to the component
                   that needs them. End with
                   the one assumption relaxed
                   vs prior work.>",
  "contribution": "(1) novel artifact,
                   (2) empirical finding /
                   design principle,
                   (3) optional dataset/tool",
  "limitation":   "<scope and applicability
                   conditions>"
}
\end{Verbatim}

  \medskip
  \textit{ANTI-PATTERN GATE} --- reject ideas whose method matches:
  \begin{enumerate}[nosep,leftmargin=*]
    \item learned independence head over tightly-coupled variables;
    \item descriptor + nearest-neighbor scheme that reduces to
          retrieval (Toolformer / Gorilla / RAG);
    \item log-prob / self-confidence as a calibrated gating signal;
    \item two cooperating sub-systems glued by a controller;
    \item ``$\langle$recent X$\rangle$ + $\langle$recent Y$\rangle$
          applied to $\langle$new domain Z$\rangle$''.
  \end{enumerate}
  For each pattern, name the specific prior method the design
  collapses onto, and name the structural property the new method
  establishes that the prior method does not.

  \medskip
  \textit{NON-OBVIOUSNESS GATE:} a domain expert reading
  \texttt{method} must NOT be able to summarize it as
  ``$\langle$named method A$\rangle$ + $\langle$named method
  B$\rangle$''. If they can, redesign.
  \end{promptbox}

  The non-obviousness gate and anti-pattern list are not part of
  the formal framework; they are guardrails added based on
  reviewer-style audits during development. Removing them yields
  the same algorithmic structure but produces noticeably more
  ``obvious composition'' ideas in our internal tests.

\section{Research Topics}
\label{sec:appendix-topics}

Table~\ref{tab:topics} lists the six research topics used in our evaluation, selected as the most frequently occurring directions in 2025 top-venue papers.

\begin{table}[h!]
  \centering
  \renewcommand{\arraystretch}{1.3}
  \resizebox{\columnwidth}{!}{%
  \begin{tabular}{@{}lp{6.5cm}@{}}
    \toprule
    \textbf{Topic} & \textbf{Description} \\
    \midrule
    \textbf{Retrieval-Augmented Generation}
      & Augmenting LLMs with external retrieval to overcome
        parametric knowledge limitations; core challenges include
        multi-hop reasoning, retrieval--generation alignment, and
        robustness to noisy context. \\
    \textbf{Agentic Reasoning}
      & Enabling LLMs to autonomously plan, use tools, and
        self-correct over multi-step tasks; key problems include
        long-horizon credit assignment, state consistency, and
        reliable reasoning under open-ended environments. \\
    \textbf{LLM Post-Training \& Adaptation}
      & Aligning pretrained LLMs with human preferences via RLHF,
        DPO, and variants; central challenges include reward
        over-optimisation, instruction-following trade-offs, and
        parameter-efficient fine-tuning expressiveness. \\
    \textbf{Multimodal LLMs}
      & Integrating vision, language, and other modalities for
        cross-modal reasoning; research focuses on fine-grained
        semantic alignment, visual hallucination mitigation, and
        spatiotemporal understanding in video. \\
    \textbf{Diffusion Models for Generation}
      & Score-based generative models for high-fidelity synthesis;
        active directions include sampling efficiency, controllable
        generation fidelity, and extending diffusion processes to
        video, 3D, and audio domains. \\
    \textbf{Efficient Inference \& Serving}
      & Maximising LLM throughput and minimising latency under
        resource constraints; core techniques span KV cache
        compression, speculative decoding, quantisation, and
        sub-quadratic attention for long contexts. \\
    \bottomrule
  \end{tabular}%
  }
  \caption{The six AI research topics used for evaluation.}
  \label{tab:topics}
\end{table}

\section{Evaluation Rubrics}
\label{sec:appendix-rubrics}

We evaluate every generated idea along five dimensions on a 10-point scale.
Table~\ref{tab:eval-dims} defines each dimension; Table~\ref{tab:rubrics} gives the full per-band rubric used by all judges.
The same rubric drives both the LLM-as-a-judge ensemble and human evaluation.

    \begin{table}[t!]
  \centering
  \renewcommand{\arraystretch}{1.3}
  \resizebox{\columnwidth}{!}{%
  \begin{tabular}{@{}lp{7cm}@{}}
    \toprule
    \textbf{Dimension} & \textbf{Definition} \\
    \midrule
    Novelty
      & Are the problems or approaches genuinely new? Is this a novel
        combination of familiar techniques? Is it clear how this idea
        differs from prior work? Is related work adequately referenced? \\
        \midrule
    Significance
      & Is the idea important? Are researchers or practitioners likely
        to use or build on it? Does it address a difficult problem in
        a better way than prior research? Does it offer a unique
        theoretical or pragmatic contribution? \\ \midrule
    Feasibility
      & Can the idea be realised with existing technology or methods?
        Are there technical bottlenecks? Is the logic sound with no
        obvious errors or unreasonable assumptions? Can a concrete
        experiment be designed from this idea? \\ \midrule
    Groundedness
      & Does every core claim---motivation, method components, and
        expected gains---trace to a specific named paper or empirical
        result? Are key assumptions supported by concrete prior work
        rather than vague ``prior work suggests'' language? \\ \midrule
    Expected Effectiveness
      & How likely is the proposed idea to outperform existing
        baselines? Is the improvement mechanism clearly articulated
        and tied to measurable outcomes? Are the performance claims
        realistic and well-scoped? \\
    \bottomrule
  \end{tabular}%
  }
  \caption{Evaluation dimensions used for both model-based and human assessment.}
  \label{tab:eval-dims}
\end{table}

\section{EvoAgent Prompts}
\label{sec:appendix-evoagent-prompt}

This section lists the prompts used by EvoAgent (\S\ref{sec:evoagent}). The
pipeline runs in three phases: \emph{Phase 1} extracts a per-tree signal
pack from $\mathcal{T}_r$ (path-level residual gaps and the cross-path join
view), \emph{Phase 2} aggregates these signals into ranked cross-path
directions, and \emph{Phase 3} converts each direction into a grounded
research idea via a type-specific draft prompt and a final novelty check.
Source: \texttt{prompts/evo\_agent\_prompts.py}.
Placeholders such as \texttt{\{topic\}}, \texttt{\{frontier\_title\}},
\texttt{\{trajectory\_ascii\}}, \texttt{\{paths\}},
\texttt{\{forest\_text\}}, \texttt{\{evidence\}}, \texttt{\{direction\}},
and \texttt{\{idea\_cards\}} are filled at run time from the toi\_tree of
the current frontier and from intermediate outputs of preceding phases.
Long structural inputs and verbatim anti-pattern lists are abbreviated as
``\dots'' for space.

\subsection{Phase 1: Path Analysis}
\label{app:prompts:phase1_path_analysis}

Run once per tree. The LLM walks every evolution path that reaches the
frontier $r$, names a structural residual gap for each path, and emits a
\emph{cross-path join view} that lists assumptions $r$ silently inherited
from \emph{multiple} parents. Source:
\texttt{get\_path\_analysis\_prompt}.

\begin{promptbox}[Path Analysis (per tree)]
You are a research evolution analysis expert performing path-level
reasoning on an evolution tree.

\medskip
\textbf{Topic:} \texttt{\{topic\}} \\
\textbf{Frontier paper (depth=0):} \texttt{\{frontier\_title\}}

\medskip
\textbf{[Tree Structure]} \texttt{\{trajectory\_ascii\}}

\medskip
\textbf{[Evolution Paths]}\quad each path: ancestor $\to$ frontier, with
its evolution action, trigger gap, mechanism change, frontier
limitation, and ancestor chain.\quad \texttt{\{paths\}}

\medskip
\textbf{[Cross-Path Join View]}\quad all mainline parents of $r$ laid
out side-by-side, with each branch's contribution, the assumption $r$
inherited from it without challenging (\texttt{what\_persists}), and
the residual gap EvoTrace already identified.\quad
\texttt{\{join\_block\_text\}}

\medskip
\textit{Task:} For each path, identify the gap that persisted across
generations without being fully resolved; then, using the join view,
identify gaps that remain unresolved across MULTIPLE branches.

\medskip
\textbf{[Structural-gap requirements]}
Every gap MUST (i) name an unestablished structural property
(an invariant, identifiability condition, closure rule, $\ldots$),
(ii) name the ancestor assumption that blocks the property,
(iii) explain why $r$'s evolution step did not close it.
Forbidden phrasings (rejected): \emph{lack of X}, \emph{no X is provided},
\emph{does not handle X}, verbatim copies of any node's limitation field,
or pure scope statements. Each gap text $\geq 40$ characters.

\medskip
Output ONLY valid JSON:
\begin{Verbatim}[fontsize=\scriptsize,breaklines=true,breakanywhere=true]
{
  "paths": [
    {
      "path_nodes": ["<ancestor>", "<frontier>"],
      "persistent_gap": {
        "gap_text": "...",
        "structural_property_name": "...",
        "ancestor_assumption_blocking_it": "...",
        "why_evolution_did_not_close_it": "..."
      },
      "why_not_resolved": "..."
    }
  ],
  "cross_path_join": {
    "branches_covered": ["A: ...", "B: ..."],
    "shared_residual_gaps": [
      {
        "gap_text": "...",
        "structural_property_name": "...",
        "branches_affected": ["A","B"],
        "common_assumption": "..."
      }
    ],
    "branch_complementarity": "..."
  }
}
\end{Verbatim}
\end{promptbox}

The \texttt{cross\_path\_join.shared\_residual\_gaps} block is the load-bearing
output: each entry pairs a structural property with the branches whose
ancestors all silently kept the assumption blocking it. These entries are
the primary evidence that Phase 2 ranks into directions.

\subsection{Phase 2: Cross-Path Direction Selection}
\label{app:prompts:phase2_aggregation}

Run once per ideation call. Given the per-tree signal packs assembled
from Phase 1, the LLM emits a ranked list of candidate research
directions. Every direction must cite at least two paths
(\texttt{evidence\_branch\_ids}, possibly across multiple trees in
\texttt{evidence\_tree\_ids}); single-path directions are rejected by the
downstream validator before idea generation. Source:
\texttt{get\_forest\_signal\_aggregation\_prompt}.

\begin{promptbox}[Cross-Path Direction Selection]
You are scanning a forest of paper evolution trees to find FUTURE research
directions.

\medskip
\textbf{Topic:} \texttt{\{topic\}} \\
\textbf{Mode:} single-tree (cross-branch evidence required) OR cross-tree
(cross-tree OR cross-branch evidence accepted).

\medskip
\textbf{[Per-tree signal pack]}\quad for each tree: residual gaps with
structural-property tags, shared residual gaps with
\texttt{branches\_affected}, recent evolution actions per branch label,
whole-tree action distribution, frontier solves, and EvoTrace v2
composite-synthesis flags.\quad \texttt{\{forest\_text\}}

\medskip
\textit{Task:} Run four signal scans, in order:
\begin{enumerate}[nosep,leftmargin=*]
  \item \texttt{convergent\_gap} --- a residual or shared residual gap
        recurs across $\geq 2$ paths.
  \item \texttt{convergent\_action} --- $\geq 2$ paths' recent actions
        point in the same direction; the future direction is the new
        bottleneck after that trajectory completes.
  \item \texttt{bridge} --- one path's \texttt{frontier\_solves} fact
        roughly matches another path's residual gap; source and target
        must be two distinct path / tree identifiers.
  \item \texttt{frontier\_race} --- $\geq 2$ recent frontiers with small
        \texttt{depth\_to\_oldest} list overlapping
        \texttt{competing\_frontiers}; the direction is a different angle
        on the shared failure mode.
\end{enumerate}

\medskip
\textbf{Absolute rules.}
Each direction MUST cite $\geq 2$ evidence units (across
\texttt{evidence\_branch\_ids} and / or \texttt{evidence\_tree\_ids}).
Reject vague directions (\emph{``improve generalization''} is not
acceptable). Reject any direction that, in one sentence, reduces to a
textbook composition of two named prior methods --- name a STRUCTURAL
PROPERTY no obvious composition can deliver, and cite the prior method
the composition would collapse onto. Never reference papers absent from
the signal pack; never fabricate a hypothetical external tree.

\medskip
Output ONLY valid JSON:
\begin{Verbatim}[fontsize=\scriptsize,breaklines=true,breakanywhere=true]
{
  "directions": [
    {
      "direction_id": "d_1",
      "type": "convergent_gap | convergent_action |
               bridge | frontier_race",
      "priority": 0.0,
      "title": "...",
      "rationale": "...",
      "evidence_tree_ids":   ["<tid>", ...],
      "evidence_branch_ids": ["<A|B|...>", ...],

      "meta_gap":         "<convergent_gap only>",
      "shared_mechanism": "<convergent_gap only>",

      "trend":                "<convergent_action only>",
      "trend_endpoint":       "<convergent_action only>",
      "implied_next_problem": "<convergent_action only>",

      "source_tree_id":   "<bridge only>",
      "target_tree_id":   "<bridge only>",
      "source_branch_id": "<bridge only>",
      "target_branch_id": "<bridge only>",
      "what_to_transfer": "<bridge only>",

      "active_battle":       "<frontier_race only>",
      "common_failure_mode": "<frontier_race only>",

      "non_obvious_property": "MANDATORY for ALL types: ..."
    }
  ]
}
\end{Verbatim}
Output directions sorted by priority (highest first). Aim for 3--8
directions; quality $>$ quantity.
\end{promptbox}

The four direction types correspond to the two cross-path configurations
described in \S\ref{sec:evoagent}: \texttt{convergent\_gap} and
\texttt{convergent\_action} instantiate the \emph{shared-gap}
configuration (the same residual property recurs across paths);
\texttt{bridge} instantiates the \emph{capability transfer} configuration
(one path supplies what another lacks); \texttt{frontier\_race} is a
multi-tree special case where multiple competing frontiers expose the
same failure mode. The \texttt{non\_obvious\_property} field is required
for every type and is the gate that prevents textbook composition from
reaching the downstream prompts.

\subsection{Phase 3: Direction-to-Idea (representative)}
\label{app:prompts:phase3_idea}

For each direction selected by Phase 2, EvoAgent dispatches a
type-specific prompt that turns the direction into a draft research idea.
All four type prompts (\texttt{convergent\_gap}, \texttt{convergent\_
action}, \texttt{bridge}, \texttt{frontier\_race}) share the same output
schema so downstream Step 3 (segment generation, reviewer / revision
loop, novelty fullcheck) can consume any of them uniformly. We include
the \texttt{convergent\_gap} prompt as a representative example; the
remaining three prompts differ only in their evidence framing and
type-specific instructions, and are otherwise identical in schema and
non-obviousness gating. Source:
\texttt{get\_convergent\_gap\_idea\_prompt} (and three siblings).

\begin{promptbox}[Direction-to-Idea (convergent\_gap)]
You are formulating a research idea for a CONVERGENT GAP --- a
mechanism-level problem unresolved across $\geq 2$ independent
paths / trees.

\medskip
\textbf{Topic:} \texttt{\{topic\}}

\medskip
\textbf{[Direction]}
\begin{itemize}[nosep,leftmargin=*]
  \item \texttt{title:}            \texttt{\{direction.title\}}
  \item \texttt{meta\_gap:}        \texttt{\{direction.meta\_gap\}}
  \item \texttt{shared\_mechanism:}
        \texttt{\{direction.shared\_mechanism\}}
  \item \texttt{priority:}         \texttt{\{direction.priority\}}
\end{itemize}

\medskip
\textbf{[Evidence pack]}\quad for each cited unit (tree or branch):
frontier title and pid, top residual gaps, top recent actions, the
direction's source / target identifiers if applicable.\quad
\texttt{\{evidence\}}

\medskip
\textit{Recurrence across independent paths is strong evidence the
underlying mechanism is a real open problem at the META level. Frame the
bridge at the structural-property level, not at any one path's
component-name level.}

\medskip
\textbf{Constraints.}
The new mechanism (\texttt{bridge\_add}) MUST address the
\texttt{meta\_gap}, not any one path's surface symptom. Convergent-gap is
the easiest type to answer with obvious composition (\emph{``everyone has
gap $G$; just bolt on module $M$''}); REJECT any \texttt{bridge\_add} a
reviewer would describe as \emph{``$\langle$named prior method$\rangle$
applied to $\langle$new domain$\rangle$''}. Before finalizing, write your
mechanism as ``$X + Y$''; if that description is accurate, redesign.
Acceptable mechanisms instead expose a STRUCTURAL PROPERTY (an
invariant, an identifiability condition, a coupling between quantities
prior work treats as independent) that no two-module composition
delivers.

\medskip
\textbf{[Anti-pattern block]}\quad five concrete anti-patterns harvested
from prior 3-judge eval failures (learned independence head over coupled
variables; descriptor + nearest-neighbor that is operationally
retrieval; log-prob as a calibrated gating signal; controller-glued
sub-systems; ``$X + Y$ in domain $Z$''). For each, the prompt names
\emph{why it fails} and \emph{how to clear the gate}.\quad \dots

\medskip
Output ONLY valid JSON:
\begin{Verbatim}[fontsize=\scriptsize,breaklines=true,breakanywhere=true]
{
  "per_trend_predictions": [
    {
      "direction":      "...",
      "rationale":      "...",
      "builds_on":      "<title of anchor paper>",
      "builds_on_pid":  "<pid present in the tree>",
      "persistent_gap": "...",
      "why_still_open": "...",
      "candidate_action": "extend|replace|migrate|merge|
                           relax_assumption|simplify",
      "bridge_keep":   "...",
      "bridge_remove": "...",
      "bridge_add":    "...",
      "non_obvious_property":
        "Property: <invariant/identity/coupling>.
         Why prior work cannot exploit it: ...
         Why composition fails: ...",
      "obvious_composition_check":
        "Shortest-form summary of the mechanism as
         X + Y; or, if composition is materially
         wrong, name the load-bearing piece."
    }
  ]
}
\end{Verbatim}
Provenance fields (\texttt{source\_trend\_ids},
\texttt{source\_branch\_ids}, \texttt{is\_cross\_tree},
\texttt{is\_cross\_branch}, \texttt{forest\_mode}) are injected by the
framework and overwritten if the LLM emits them.
\end{promptbox}

After Phase 3 produces a draft prediction, the prediction is fed into a
shared idea-generation chain that
performs segment-level drafting, an adversarial reviewer/revision loop,
and a final novelty fullcheck against the retained tree. The remaining three
direction-type prompts (\texttt{convergent\_action}, \texttt{bridge},
\texttt{frontier\_race}) follow the same skeleton: a type-specific
``why this is a hard bone'' framing block, the same anti-pattern list,
and the same JSON output schema.

\section{Evaluation Prompt}
\label{sec:appendix-eval-prompt}

The following prompt is used for model-based evaluation.
\texttt{\{topic\}} and \texttt{\{idea\}} are filled with the
research topic and the idea under evaluation, respectively.
The five rubric blocks are identical to those in
Table~\ref{tab:rubrics} and are inlined into the prompt at
run time.

\begin{promptbox}[Main Evaluation Prompt]
You are an extremely demanding scientific reviewer with the highest
critical standards, like those at top venues (Nature, Science, ICML,
NeurIPS). Most submitted ideas have fundamental flaws and should not
exceed 6 in any dimension. Reserve 7+ for ideas with concrete,
demonstrable strengths; reserve 9--10 for ideas that would
meaningfully shift the field.

\medskip
\textbf{TOPIC:} \texttt{\{topic\}}

\medskip
\textbf{IDEA:} \texttt{\{idea\}}

\medskip
Evaluate the idea using the five rubrics below. Assess intrinsic
quality, not writing polish. Focus on whether the idea would be
worth pursuing as a research project.

\medskip
\textit{[Five rubric blocks inlined here --- identical to
Table~\ref{tab:rubrics}.]}

\medskip
\textit{Evaluation procedure:}
\begin{enumerate}[nosep,leftmargin=*]
  \item \textbf{Novelty only} --- Before scoring, internally identify
    2--3 existing works most similar to this idea and the core
    mechanism difference for each. Do \emph{not} output this
    analysis; use it only to ground your novelty score.
  \item For each dimension, internally identify the strongest
    evidence for and against.
  \item Map your judgment explicitly to a rubric band, then assign
    an integer score 1--10.
\end{enumerate}

\medskip
\textit{Scale calibration (apply to every evaluation):}
\begin{itemize}[nosep,leftmargin=*]
  \item Use the full 1--10 range. An excellent idea earns 9--10; a
    fundamentally flawed one earns 1--2. Clustering all scores in
    6--8 is a sign of miscalibration.
  \item Be a critical reviewer. An idea that merely ``sounds
    reasonable'' is at most 6. To exceed 6 in any dimension, the
    idea must demonstrate a specific concrete strength matching the
    band description.
\end{itemize}

\medskip
\textit{Per-dimension guidance:}
\begin{itemize}[nosep,leftmargin=*]
  \item \textbf{Novelty}: complete the prior-work comparison before
    scoring. Trivial combinations score 5--6; non-obvious
    combinations that a domain expert would \emph{not} have
    anticipated score 7--8.
  \item \textbf{Significance}: evaluate the solution's impact, not
    just the problem's importance.
  \item \textbf{Groundedness}: each key claim must be traceable to a
    named reference or stated logical derivation.
  \item \textbf{Feasibility}: score [A]~technical realizability and
    [B]~experiment design completeness internally, then return
    Final $= \mathrm{round}((A+B)/2)$. A technically broken method
    cannot score above 6.
  \item \textbf{Effectiveness}: evaluate whether the causal chain
    from method to claimed gain is complete.
\end{itemize}

\medskip
Return \textbf{only} one valid JSON object with exactly these six
fields and no other keys. The \texttt{brief\_analysis} field must be
$\leq$60 words and must (a) name the 2 closest prior works (or say
``no close prior work identified'' if genuinely none), and (b) state
the single biggest weakness. Do not include markdown fences,
reasoning outside the JSON, or any other text.

\begin{Verbatim}[fontsize=\scriptsize,breaklines=true,breakanywhere=true]
{
  "brief_analysis": "<= 60 words: 2 closest prior works
                      + single biggest weakness",
  "novelty":        1-10,
  "significance":   1-10,
  "groundedness":   1-10,
  "feasibility":    1-10,
  "effectiveness":  1-10
}
\end{Verbatim}
\end{promptbox}

\balance

\begin{table*}[t]
  \centering\small
  \renewcommand{\arraystretch}{1.2} 
  \caption{Scoring rubrics for all five evaluation dimensions.
    For Feasibility, [A] and [B] are scored independently;
    final score $= \min(A,B)$ if $A < 5$, else $\mathrm{round}(0.6A + 0.4B)$;
    if $A \leq 3$ the final score cannot exceed $A$.
    If $A \leq 6$, [B] is capped at 7 unless a validation scheme targeting
    the specific technical hurdle is present.}
  \label{tab:rubrics}
  \begin{tabularx}{\textwidth}{@{}lX@{}} 
    
    \midrule
    \multicolumn{2}{@{}l@{}}{\textbf{Dimension: Novelty}} \\
    \midrule
    9--10 & The problem formulation or the core mechanism is itself new (no prior work has proposed it independently); not merely a new combination of two existing ideas; related work thoroughly covered. \\
    7--8 & Both problem and mechanism have individual precedents, but their combination is non-obvious to a domain expert and has not been demonstrated before; structural difference from closest prior work explicitly stated; no obvious related work omissions. \\
    5--6 & Novel combination of known components; combination is obvious in hindsight — a domain expert would have anticipated it without seeing this work. \\
    3--4 & Minor variant (hyperparameters, encoder swap, layer count); domain transfer with no method adaptation; no named prior work identified to contrast against. \\
    1--2 & Restates or relabels an existing method; core idea already present in a top-venue paper. \\
    
    \midrule
    \multicolumn{2}{@{}l@{}}{\textbf{Dimension: Significance}} \\
    \midrule
    9--10 & Addresses a core bottleneck limiting multiple research threads; solution specific enough to resolve it; contribution is citable and buildable-upon by others. \\
    7--8 & Solves a real important problem in an identifiable sub-community; significance of the solution is argued (not just asserted); at least one downstream use case identified. \\
    5--6 & Addresses a real but narrow problem; impact plausibly limited to the specific setup; solution's generalizability unclear. \\
    3--4 & Improvement marginal over existing solutions; OR solution too vague to deliver the claimed impact regardless of problem importance. \\
    1--2 & Problem synthetic or contrived; even if fully realized, the broader community would not notice. \\
    
    \midrule
    \multicolumn{2}{@{}l@{}}{\textbf{Dimension: Feasibility}} \\
    \midrule
    9--10 & All components (models, data, APIs) exist and are accessible; no unresolved algorithmic breakthroughs needed; completable within one conference cycle under routine resources. [B]~Datasets, baselines, and primary metrics are stated or directly inferable; at least one ablation or alternative validation scheme can be reasonably anticipated. \\
    7--8 & One moderate technical challenge exists, but a known, documented solution or mature engineering practice is available; no original engineering research required to overcome it. [B]~Most validation elements specified; only one gap remains with narrow, well-understood choices. \\
    5--6 & A significant technical hurdle exists but falls within an established theoretical path; authors do not state a solution and domain experts cannot easily infer one. [B]~Outline present but 2+ key elements missing. \\
    3--4 & Requires core capabilities that do not currently exist and for which no clear roadmap exists in the foreseeable future (3--5 years). [B]~No evaluation protocol described or inferable. \\
    1--2 & Fundamentally unrealizable with current technology, or the method contains an internal technical contradiction. [B]~No path to empirical validation exists whatsoever. \\
    
    \midrule
    \multicolumn{2}{@{}l@{}}{\textbf{Dimension: Groundedness}} \\
    \midrule
    9--10 & Every claim (problem exists, gap is real, design choice) anchored to a named paper or known empirical fact; no ``prior work suggests'' language anywhere. \\
    7--8 & Primary claims backed by named literature; minor details may lack citations but no core claim floats free. \\
    5--6 & Some named citations present, but at least one key claim relies on unnamed ``prior work'' or field intuition. \\
    3--4 & Motivation generic (``existing methods have limitations'') with no named failure case; method choices asserted without any traceable source. \\
    1--2 & No named prior work anywhere; every claim asserted from scratch. \\
    
    \midrule
    \multicolumn{2}{@{}l@{}}{\textbf{Dimension: Effectiveness}} \\
    \midrule
    9--10 & Improvement tied to a specific baseline failure mode via an explicit causal argument (``X fails because Y; we address Y by introducing W''); expected gains name a specific metric and direction; no ``SOTA on all'' claims. \\
    7--8 & Improvement mechanism stated and plausible; link to a baseline weakness is inferable even if not spelled out; performance claims name a metric and direction. \\
    5--6 & Method described but why it outperforms baseline is hand-wavy (``more flexible/general/expressive''); OR causal chain has one gap the reader must fill with their own assumptions. \\
    3--4 & No mechanism connecting method to claimed gain; improvement simply asserted; OR method addresses a different problem than stated in motivation. \\
    1--2 & Method cannot logically address the stated problem even in principle; causal chain absent or self-contradictory. \\
    
    \bottomrule
  \end{tabularx}
\end{table*}
\end{document}